 \documentclass[final,5p,times,twocolumn]{elsarticle}

\usepackage{graphicx}
\usepackage{xcolor} 
\usepackage{amssymb}

\usepackage{amsthm}     
\usepackage{subfigure}

\usepackage{float}

\usepackage{hyperref}
\usepackage{xspace}  

\usepackage{multirow}

\journal{.}
\begin{document}
\begin{frontmatter}


\title{Spatiotemporal deep learning models for  detection of rapid intensification in cyclones}


\author[iit,pingla,]{Vamshika Sutar}

\author[iit]{Amandeep Singh}

\author[unsw2]{Rohitash Chandra} 
\ead{rohitash.chandra@unsw.edu.au} 
  
\address[iit]{Department of Civil Engineering, Indian Institute of Technology Bombay, Mumbai,  India}

\address[pingla]{Centre for Artificial Intelligence and Innovation, Pingla Institute, Sydney, Australia} 
 
\address[unsw2]{Transitional Artificial Intelligence Research Group, School of Mathematics and Statistics, University of New South Wales,  Sydney,  Australia}

\begin{abstract}
  
Cyclone rapid intensification is the rapid increase in cyclone wind intensity, exceeding a threshold of 30 knots, within 24 hours. Rapid intensification is considered an extreme event during a cyclone, and its occurrence is relatively rare, contributing to a class imbalance in the dataset. A diverse array of factors influences the likelihood of a cyclone undergoing rapid intensification, further complicating the task for conventional machine learning models. In this paper, we evaluate deep learning, ensemble learning and data augmentation frameworks to detect cyclone rapid intensification based on wind intensity and spatial coordinates. We note that conventional data augmentation methods cannot be utilised for generating spatiotemporal patterns replicating cyclones that undergo rapid intensification. Therefore, our framework employs deep learning models to generate spatial coordinates and wind intensity that replicate cyclones to address the class imbalance problem of rapid intensification. We also use a deep learning model for the classification module within the data augmentation framework to differentiate between rapid and non-rapid intensification events during a cyclone. Our results show that   data augmentation improves the results for rapid intensification detection in cyclones, and spatial coordinates play a critical role as input features to the given models.  This paves the way for research in synthetic data generation for spatiotemporal data with extreme events. 


\end{abstract}

\begin{keyword}
Cyclones \sep class imbalance problem \sep deep learning \sep rapid intensification.

\end{keyword}

\end{frontmatter}




\section{Introduction}
\label{S:1} 
 
 Over the past decade, the impacts of climate change have manifested in an alarming increase in the strength of tropical cyclones, characterised by elevated levels of precipitation and wind intensity, resulting in devastating consequences on a global scale \citep{peduzzi2012global,miyamoto2015triggering,wehner2019estimating}.    Rappaport et al.  \citep{rappaport2012joint} defined rapid intensification as a sudden surge in wind intensity exceeding 30 knots (35 miles/hour or 55 kilometres/hour) within 24 hours \citep{nhc2013glossary}. Forecasting the rapid intensification of high-category cyclones (Category 4 and 5) poses greater challenges due to their infrequent occurrence, in contrast to lower-category cyclones\citep{bhatia2019recent}. Due to limited data, there is much speculation about the exact cause of rapid intensification; some of the potential causes are changes in the oceanic-atmospheric conditions \citep{kaplan2015evaluating}, vertical wind shear from the upper-level trough \citep{molinari1990external}, elevated levels of oceanic heat \citep{kaplan2010revised}, insufficient cooling of inner-core area Sea Surface Temperature (SST) \citep{cione2003sea}, and location of the cyclone vortex near the radius of maximum wind \citep{rogers2015multiscale}.

 The utilisation of deep learning methods for cyclone intensity prediction is an evolving and emerging field \citep{chen2020machine}. Similarly, cyclone genesis \citep{chen2019hybrid}, track prediction \citep{lian2020novel}, and intensity prediction \citep{yuan2021typhoon} have been implemented in deep learning models such as  Long Short-Term Memory  (LSTM) recurrent networks \cite{hochreiter1997long}.   \citet{wang2020tropical} used deep learning via convolutional neural networks (CNNs)  to predict tropical cyclone intensity using ocean-atmospheric data as features.   \citet{varalakshmi2021tropical} used CNNs to categorise cyclone wind intensity in terms of the level of destruction.   \citet{wimmers2019using} used CNNs to predict the cyclone wind category using satellite image data. Furthermore, several studies utilised  CNNs to estimate cyclone intensity (category) using satellite imagery and cloud pattern data  \citep{maskey2020deepti,wimmers2019using,rajesh2020prediction,lee2020tropical}. 

 
    \citet{maskey2020deepti} used  CNNs with satellite data to predict cyclone intensity and reported consistency between the regions of data (pixels) with high correlation, which is typically observed by experts to make the decision.  \citet{lee2020tropical}   used deep learning to understand the inherent cyclone characteristics and categorised cyclone intensity using multi-spectral satellite data. The authors reported that a higher-intensity cyclone influences the strength of the cyclone centre, exerting a more pronounced effect in the lower atmosphere. The accuracy of deep learning methods has demonstrated results relatively comparable to numerical weather models, suggesting their potential use for estimations at arbitrary cyclone positions \citep{wei2019study}.    \citet{cloud2019feed} demonstrated that a simple neural network (multilayer perceptron) outperformed a hurricane weather forecasting model. In terms of using existing numeral data (geolocation and wind-intensity);   \citet{Chandra-Dayal2015,chandra2016architecture} used coevolutionary recurrent neural networks (RNNs) for cyclone track (path) prediction, and later   \citet{ZhangChandra2018} used matrix neural networks for cyclone path prediction which showed better performance than conventional deep learning models.  \citet{ChandraOG18-asoc} used coevolutionary multitask learning for dynamic cyclone wind-intensity prediction \citep{Chandra2017-iconip} with the motivation to make a prediction as soon as possible, given limited data about the cyclone. Furthermore,   \citet{Kapoor2023} used variational RNN and LSTM models to provide uncertainty quantification with the prediction of cyclone tracks and wind intensity.  These studies motivate the use of deep learning methods for cyclone modelling and detection of rapid intensification events.

   Several studies have been done to understand the difference between a rapid intensification (RI) and a non-RI event, to improve statistical and machine learning models.   \citet{fischer2019climatological} studied the interaction between tropical cyclones and upper-tropospheric troughs and related it to  RI events  \citep{lu2021relationship}.   \citet{kaplan2015evaluating} presented a study that featured statistical models such as logistic regression for predicting/detecting rapid intensification, which demonstrated poorer accuracy when compared to an established numerical model (Hurricane Weather Research and Forecasting model) \citep{kaplan2015evaluating}. The authors reported class imbalance in detecting rapid intensification; however, there was no attempt to address the class imbalance via deep data augmentation methods.   \citet{su2020applying} demonstrated the correlation of internal storm structure with rapid intensification, which was then used as a feature for the machine learning model.   \citet{yang2020long} used  LSTM model to predict rapid intensification based on cyclone (storm) features, and related environmental and basin features.
Coevolutionary RNNs \citep{chandra2015coevolutionary,chandra2017towards} have been used for rapid intensification detection based on wind-intensity, which highlighted the challenge of class imbalance.


The challenge of class imbalance in data has been a focal point in various research domains, and not much has been done in the area of data augmentation for spatiotemporal problems. We note that the Synthetic Minority Oversampling Technique (SMOTE) \cite{chawla2002smote}
for regression \cite{torgo2013smote} can be used to address time series problems; however, it does not have the inbuilt mechanism to account for the spatial nature of the data. In particular, our RI detection is a class imbalance problem arising from the spatiotemporal problem, and to generate data, we need to ensure that the methodology has the potential to generate spatial coordinates that represent tracks of cyclones undergoing RI. This generation of tracks as images can be resolved by Generative Adversarial Networks (GANs) \cite{goodfellow_generative_2014} if we process the tracks as images; however, we also need wind intensity to be generated along the tracks.  Therefore, unlike traditional data augmentation techniques, we can leverage the capabilities of LSTMs to generate synthetic samples.

  In this paper, we evaluate deep learning, ensemble learning and data augmentation frameworks LSTMs to detect rapid intensification of cyclones. We note that conventional data augmentation methods cannot be utilised to generate spatiotemporal patterns replicating cyclones undergoing rapid intensification. Therefore, our framework employs deep learning models to generate (spatial coordinates and wind intensity) that replicate cyclones to address the class imbalance problem of rapid intensification. This process involves learning the underlying patterns of the existing RI cyclone data to generate synthetic samples that enhance the representation of the minority class (RI cases).  We also use a deep learning model for the classification module within the data augmentation framework to differentiate between rapid and non-rapid intensification events during a cyclone. Our methodology aims to overcome the limitations associated with local data distribution and noise often encountered in traditional techniques.  Furthermore, we evaluate a wide range of models that include  1.) univariate, 2.) multivariate, 3.) LSTM networks combined using ensemble learning, and 4.) multivariate models combined with data augmentation. 
We also introduce a data augmentation framework that utilises the LSTM model to detect cyclone RI events based on wind intensity and the cyclone's spatial coordinates. Our framework features a data augmentation component that enables an  LSTM model to generate wind intensities and spatial coordinates of cyclones based on cyclones that had RI events. This process involves learning the underlying patterns of the existing RI cyclone data to generate synthetic samples that enhance the representation of the minority class (RI cases).
Therefore, we contribute to the broader understanding of data augmentation techniques for improving model performance in scenarios with imbalanced classes arising from spatiotemporal data.


The rest of the paper is organised as follows.   Section 2  presents the methodology that provides an evaluation of deep learning models and data augmentation framework for the identification of cyclone rapid intensification. Section 3 presents the results which is followed by discussion in Section 4, and finally, Section 6 concludes the paper.

\section{Related Work } 

\subsection{Class imbalance}

 Supervised model training generally shows a bias towards the majority class. Class imbalance problems are naturally present in many real-world problems \citep{buda2018systematic,
galar2011review,abd2013review,gosain2017handling}, where one class would have significantly more data than the other, which also applies to multi-class classification problems. The class imbalance problem is present in cyclone RI \citep{wei2021advanced,chand2017projected}, which comes with several challenges for modeling.  There are further challenges with multi-class classification, high-dimensional data, and hierarchical order of subcategories in data \citep{raghuwanshi2020smote}. In certain real-world applications,  the minority class is of interest; for instance, when predicting climate extreme events, where the “normal" class would be higher than the extreme event, such as cyclone RI.  In the area of cyclones, significant upward trends for wind intensity have been found \citep{elsner2008increasing}, and often wind intensity is used to classify the tropical cyclone categories. High-intensity cyclones are becoming more prevalent \citep{bhatia2019recent}; however, the RI cyclone events are still intrinsically imbalanced in the data. Class imbalance problem can be addressed with novel strategies at the data processing stage through data augmentation \citep{johnson2019survey}, and at the model development stage \citep{elsner2008increasing}. Over-sampling of the minority class and under-sampling of the majority class have been commonly used in machine learning to address class imbalance problems. Under-sampling in data augmentation may result in loss of information and reduce the class representation in the dataset \citep{sitompul2018biased}; most of the literature recommends oversampling \cite{xu2019cyclone}. At the model development stage, specific algorithms can be used so that the model architecture can adjust the learning process, thereby increasing the importance of the positive class. Innovative machine learning methods \citep{yan2018classifying,wang2012applying,Khan2024} can use a combination of both techniques by data sampling and applying cost-sensitive learning to reduce bias.


Data augmentation methods have been used to address class imbalance problems. 
  SMOTE  \citep{chawla2002smote,raghuwanshi2020smote} has been prominent in addressing the class imbalance problem, where synthetic minority class samples (data) are created by interpolating between existing minority samples. In this way,  the creation of minority samples balance the majority classes. Although a widely used technique, SMOTE has limitations \citep{fernandez2018smote} since it only takes into account local data distribution, which results in noisy minority samples that are not well-matched with the global data distribution, and variants have been proposed \citep{wang2020global}. GANs  \citep{goodfellow_generative_2014} provide another data augmentation approach to address class imbalance problems \citep{ali2019mfc,mariani2018bagan,sharma2021smotified}.   GAN consists of two neural network models, i.e.,  generator and discriminator networks that oppose each other but are concurrently trained in the form of a zero-sum game, where one network's gain is the loss for the other.   GANs have been mostly used for computer vision-related tasks \citep{creswell2018generative,gui2021review} and applications,  such as medical image analysis \citep{yi2019generative}. The use of GANs for addressing class imbalance problems is gaining attention, both for image-based and tabular datasets \cite{sharma2021smotified,scott2019gan,nafi2020addressing}. A comprehensive evaluation has also been done on the combination of data augmentation methods such as SMOTE and GAN with ensemble learning for class imbalance problems \citep{Khan2024}.   These studies motivate the use of data augmentation for the detection of cyclone RI events. 


 \subsection{Cyclone analysis}

 Climate projections suggest an increase in cyclone intensity by approximately 10 percent, with a related increase in cyclone induced precipitation rate by 20 percent as we approach 2100~\citep{knutson2010tropical}. However, the average cyclone frequency is projected to incur decrease, which can highly vary based on the cyclone basin ~\citep{murakami2020detected}. Tauvale et al. \citep{tauvale2019characteristics} presented a recent study on characteristics of cyclones in the South West Pacific and reported a decrease in the annual number of cyclone duration (days) and an increase in the numbers of higher intensity cyclones  ~\citep{tauvale2019characteristics}. The historical variability trends of cyclones indicate that a more active successive season follows a period of low cyclonic activity; therefore, the decrease in cyclone frequency should not be discounted ~\citep{magee2019historical}. The El Nino Southern Oscillation (ENSO) influences cyclone genesis ~\citep{magee2017influence}, and climate projections have shown that ENSO-driven variability has shown an increase in cyclone frequency by approximately 20-40 percent ~\citep{chand2017projected}. In addition, ENSO has also been found to influence the track of cyclones where it has been reported that La Niña seasons drove cyclones to take a more southerly track, whereas the El Nino season induced more frequent north-westerly tracks ~\citep{mcinnes2011progress}. Cyclones cause floods of higher magnitude ~\citep{kostaschuk2001tropical,pradhan2007risk}, based on factors such as cyclone velocity, intensity, and duration ~\citep{madsen2004cyclone,terry2009emerging}. It is crucial to consider the cascading effects of cyclones, which can cause a series of events at the same temporal and spatial scale namely, flooding, cyclone and associated storm surge, which significantly increases the magnitude of a disaster event ~\citep{alexander2018magnitude}. There are only six cyclone basins, where the  Western Pacific Basin is the most active  ~\citep{munshi2015general}.

  \section{Methodology}
 
 \subsection{Data and study region}

\begin{figure*}[htbp]
\centering
\includegraphics[width = \linewidth]{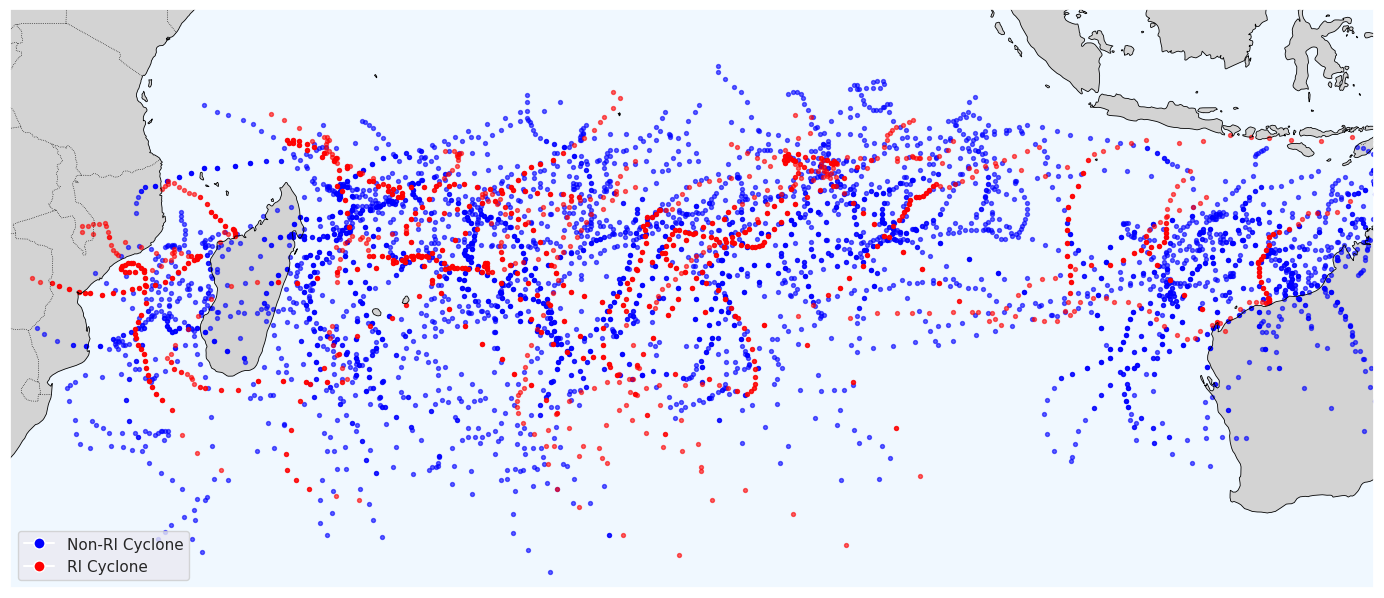}
\caption{Spatio-temporal tracks of cyclones in the South Indian Ocean basin from 2009 to 2019, highlighting instances of Rapid Intensification (RI) and non-RI
events.}

\label{fig:SI_RI/NONRI}
\end{figure*}

In this study,   we consider the Southern Hemisphere, South Pacific and South-West Indian Ocean as the basins of interest (Figure \ref{fig:SI_RI/NONRI}). The equatorial region in the Southern Hemisphere provides optimum conditions for the formation of cyclones in the region \citep{steenkamp2019tropical}. The South-East Indian (SEI) Ocean spans from 90 degrees East onwards, where part of the region is monitored by the Australian Bureau of Meteorology. The Southern Pacific (SP) region spans from 25 degrees South, 160 degrees East – 120 degrees West. This region is governed by World meteorological official centres, mainly the Fiji Meteorological Service, and the Meteorological Services of New Zealand. The cyclone data was extracted from the Joint Typhoon Warning Centre (JTWC) website \footnote{\url{https://www.metoc.navy.mil/jtwc/jtwc.html?best-tracks}} \citep{JWTC}, which presents it as \textit{best track} cyclone data. The source contains location in terms of cyclone track, and cyclone intensity given in knots (i.e., the maximum 1-minute mean sustained 10-meter wind speed, where 1 knot equals 1.852 kilometre per hour)  at six-hour intervals. We found missing data for the cyclones in the earlier years; hence, we used data from 1980-2020, as shown in Table \ref{tab:data}. 
 
\begin{figure*}[htbp]
\centering
\includegraphics[width = \linewidth]{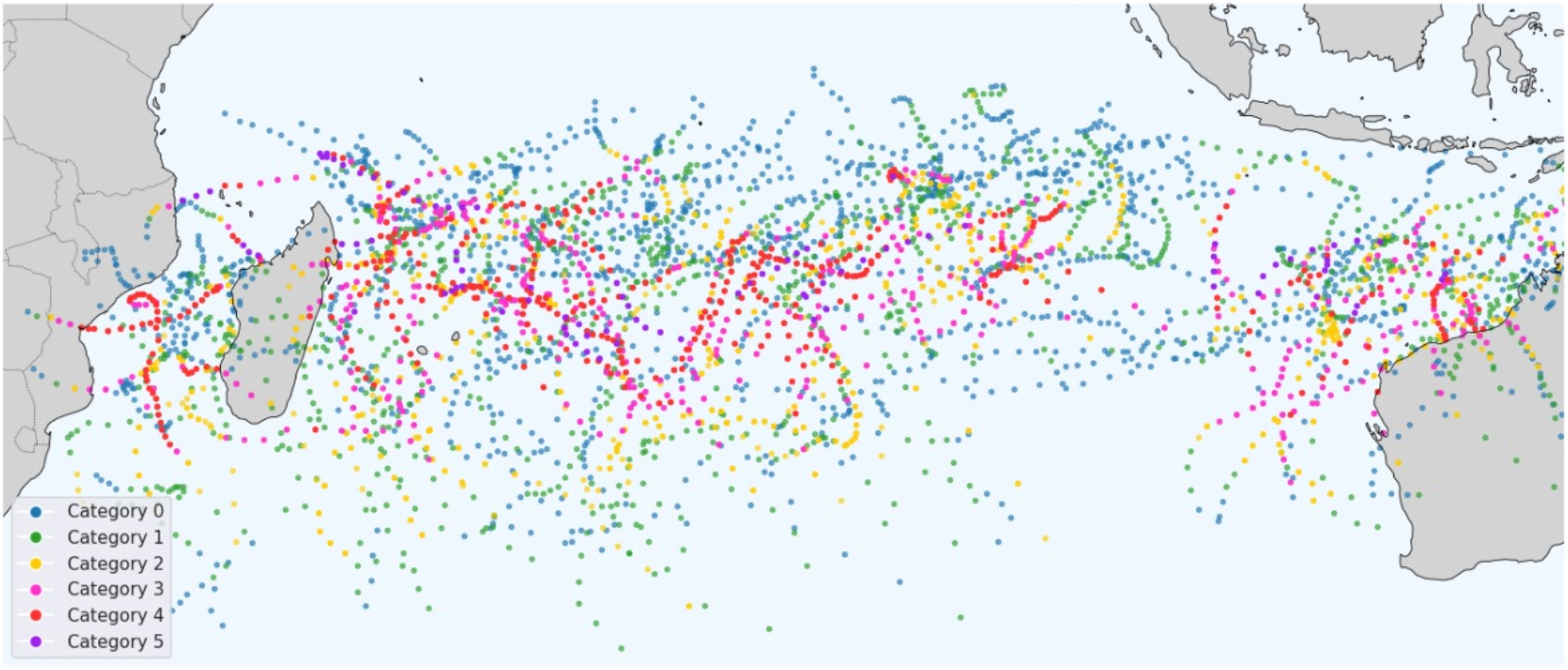}
\caption{Spatio-temporal tracks of cyclones in the South Indian Ocean basin, categorized by wind speed-based classifications.}

\label{fig:SI_categories}
\end{figure*}

 \subsection{Deep learning models}
 
  Deep learning \citep{lecun2015deep} is a prominent field of machine learning that has been prominent for big data and computer vision  \citep{guo2016deep} and is slowly becoming prominent in climate and Earth sciences \citep{scher2018toward,ardabili2019deep}. Although deep learning methods have outperformed conventional machine learning methods \citep{wang2021comparative}, they lack interpretability and hence have been the focus of explainable machine learning \citep{roscher2020explainable}. However, in many applications,  the primary goal has been to improve models that give better predictions, and the goal of having explainable models has been secondary.  RNNs are prominent deep learning methods suited for modelling temporal sequences \citep{hochreiter1997long} that are derived from feedforward neural networks with a major difference being feedback of information which is typically implemented using a context layer   \citep{elman1990finding}. In this way, RNNs exhibit temporal dynamic behaviour and have been proven to be equivalent to certain types of dynamical systems \citep{giles1999equivalence}.  RNN training has been done using the backpropagation through-time algorithm developed by Werbos \citep{werbos1990backpropagation}, which is an extension of backpropagation. RNNs had problems in representing and training from temporal sequences with long-term dependencies, resulting in the vanishing gradient problem during training \citep{bengio1994learning}. The LSTM networks addressed the vanishing gradient problem with augmented memory structures that retained information for sequences with long-term dependencies \citep{bai2019ensemble,mikolov2014learning}.  

 LSTM models have been prominent in a wide range of applications involving temporal sequences, natural language processing and computer vision   \citep{van2020review}.   LSTM models have been used for spatiotemporal problems \citep{shi2018machine} and climate extremes, such as hydrological forecasting \citep{ding2020interpretable} and rainfall prediction \citep{xu2019preliminary}.  LSTM models can be easily combined with CNNs \citep{ming2020survey},  and convolutional layers from CNNs have been incorporated in LSTM implementations \citep{agrawal2021long} with promising results for spatiotemporal problems. The LSTM models and their variations, such as encoder-decoder LSTM and bidirectional LSTM, also suit multi-step ahead time series prediction;  however, a study showed that they perform slightly poorer when compared to CNNs but better than shallow neural networks \citep{chandra2021evaluation}. 

 We use the LSTM model as the foundational model for evaluating univariate and multivariate strategies, and also for data augmentation and the classification component for RI events. Specifically, we employ an LSTM model to generate synthetic data for the minority class of RI events, aligning them with non-RI events to mitigate class imbalance. 
 
\subsection{Spatiotemporal data augmentation framework}

 The primary contribution of our study lies in the development and evaluation of conventional deep learning and data augmentation-powered deep learning models using spatio-temporal data for addressing the challenges posed by class imbalance. We evaluate the respective models for spatiotemporal rapid intensification classification problems with class imbalance featuring cyclones spanning the last few decades from selected ocean basins, namely South Pacific and South-West Indian Ocean (Figure \ref{fig:SI_RI/NONRI}). 

\subsubsection{Deep learning strategies}

\begin{figure*}[htbp]
\centering
\includegraphics[width=17cm]{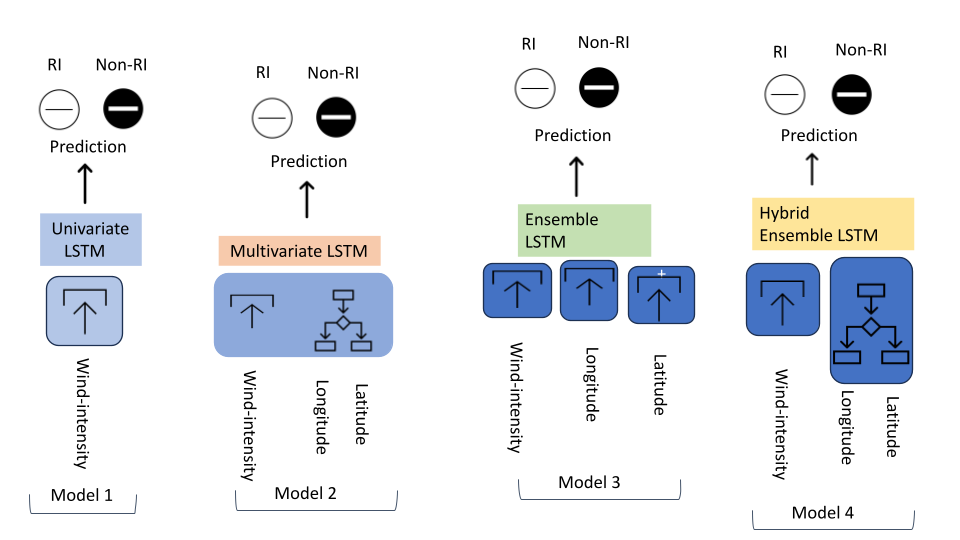}
\caption{LSTM model configurations for predicting rapid intensification of cyclones.}
\label{fig:models}
\end{figure*}

\begin{figure*}[htbp]
\centering
\includegraphics[width=17cm]{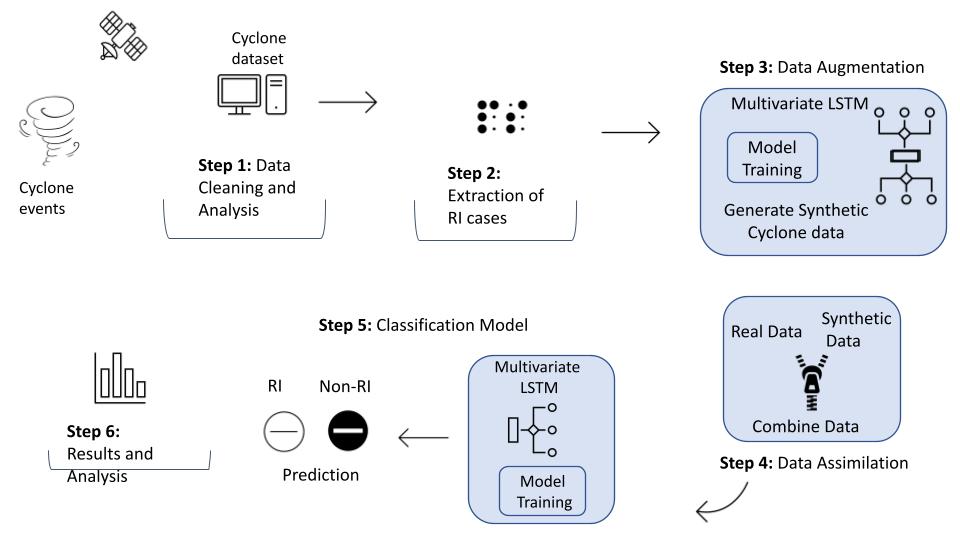}
\caption{Proposed framework for cyclone RI prediction: The process involves data cleaning and RI case extraction, followed by LSTM-based data augmentation
(Step 3), data assimilation (Step 4), and final classification using a multivariate LSTM model (Step 5)}
\label{fig:framework}
\end{figure*}

 In this study,  we
evaluate a wide range ofLong Short Term Memory (LSTM) models (Figure \ref{fig:models}) that include
1.) Univariate LSTM, 2.) Multivariate LSTM, and 3.) Ensemble LSTM, 4.) Hybrid Ensemble LSTM 5.) Data Augmentation Multivariate LSTM framework (Figure \ref{fig:framework}) for
the detection of RI events. We employ LSTM models both for the detection of RI events and as a data augmentation tool to generate spatiotemporal data for an over-sampling strategy to address class imbalance.   The  LSTM model for data argumentation ensures that the data generated is representative of the original data with underlying patterns in the spatiotemporal data.  We train the  Univariate LSTM   (U-LSTM) model exclusively on wind intensity time series data. Our goal is to investigate the model's ability to capture temporal dependencies in isolation when trained on only wind intensity. The Multivariate LSTM (M-LSTM) model incorporates both wind intensity and spatiotemporal coordinates (latitude and longitude) as input features. This model intends to explore potential improvements in prediction accuracy by considering additional environmental contexts. The Ensemble LSTM (E-LSTM) consists of three separate Univariate LSTM models. We train each LSTM in the ensemble using one of three input features: wind intensity, latitude, and longitude. E-LSTM leverages the strengths of  Univariate LSTM models and investigates the efficacy of their combined strength. 
Finally, the Hybrid-Ensemble LSTM (HE-LSTM) comprises a Univariate LSTM trained on wind intensity and a multivariate LSTM trained on spatiotemporal coordinates. This hybrid model combines univariate and multivariate information. The diverse set of LSTM models enables a comprehensive analysis of the input features on the models' predictive performance.

\begin{figure}[htbp]  
\centering
\includegraphics[width=0.9\linewidth]{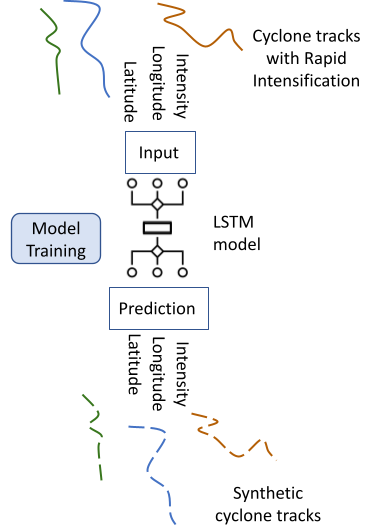}
\caption{Illustration of the data augmentation strategy using an LSTM model. This highlights Step 3 of the proposed framework (see Figure~\ref{fig:framework}}
\label{fig:generate}
\end{figure}

 \subsubsection{Data augmentation framework}
  
  We note that RI events are rare; hence, we have a class imbalance problem that can be addressed by generating synthetic data through novel data augmentation strategies. However, the majority of such strategies in the literature have not addressed data augmentation in the case of spatiotemporal sequences. The focus has been on data augmentation for text and image datasets \citep{shorten2021text,shorten2019survey} and limited work has been done in data augmentation in the case of temporal data \citep{wen2020time}. Our data augmentation strategy utilises the existing cyclones with RI events from the respective cyclone datasets (defined by ocean basins) by using the spatial coordinates and wind intensity attributes (Figure \ref{fig:generate}). We select the LSTM model for its inherent capability to capture temporal dependencies and employ it as a generator of spatiotemporal cyclone data.  Figure \ref{fig:framework} presents our framework that indicates how the data is taken from the source, cleaned and processed. We employ the data augmentation (oversampling) component to generate the synthetic data (Step 3) that is later combined with existing real data (Step 4) for the detection of RI events. Our cyclone dataset features cyclones with latitude, longitude and wind intensity (knots); every data point is recorded at six-hour intervals. 

In the LSTM model used for data augmentation,  we use a 4-time step model input (representing the past 24 hours)   to predict the subsequent 4-time steps based on the evolving context of each cyclone. Therefore, we feature a multivariate model that receives three features (latitude, longitude, wind intensity) of 4 time steps in the past, for the prediction of future four time steps (Figure \ref{fig:framework}.  This architectural choice ensured that the LSTM learns and encapsulates the temporal intricacies governing cyclonic phenomena. The model's grasp of temporal dynamics, coupled with its contextual understanding, can potentially incorporate realistic fluctuations and variations into the spatial coordinates and wind intensity trajectories of cyclones.

 During the training phase (Figure \ref{fig:framework} -- Step 3), the LSTM features a comprehensive dataset that includes instances of both rapid intensification (RI) and non-RI cyclones (Figure \ref{fig:generate}),and in the test set, we had only RI cyclones, so that later we can add the newly generated 4 points to the original 4 data points of RI instances, later added to train set of the original dataset, for training the classification model. This is done to enable the LSTM model to develop a robust understanding of the diverse range of cyclonic patterns present in both RI and non-RI instances.
 
Although not a requirement, to enhance our understanding, we also evaluate the quality of generated data by using a test set of actual RI cases (Figure \ref{fig:framework} -- Step 3).   This strategy aligns with the overarching goal of our study, which is to augment quality data specifically for the minority class, i.e., RI events in cyclones, which are often underrepresented due to their infrequent occurrence. It is well known that poor-quality data samples would further deteriorate the model for the detection of RI events. Hence, it is important to validate the quality of the samples generated. 

We finally augment  LSTM model predictions (synthetic data) encompassing the spatial coordinates and wind intensities associated with RI cyclones (Figure \ref{fig:framework}--Step 4).  This process mitigates the class imbalance issue endemic to imbalanced datasets, where the scarcity of RI events can impede the model's ability to generalise effectively. In the final stage, we train the RI detection module (Step 5) that features another LSTM model with the augmented dataset that features real and synthetic RI events. We utilise a multivariate LSTM model that takes advantage of the spatio-temporal nature of the dataset.

 \subsection{Metrics}

 We note that our focus is on class imbalance; hence, we report a wide range of metrics to ensure we effectively quantify the predictions. The metrics using the predicted outcomes, such as  true positive (TP), false positive (FP), true negative (TN), and false negative (FN) given below:

 
    \begin{equation}
Precision = \frac{TP}{TP+FP}
\end{equation}

    \begin{equation}
Recall = \frac{TP}{TP+FN}
\end{equation}
\
\begin{equation}
    F1  = \frac{2*Precision*Recall}{Precision+Recall}
    \end{equation}


The macro-average ($\hat{m}$) precision and recall is the average of precision (P) and recall (R) metrics for all the individual classes. In the case of 2 classes, the  precision is denoted by $P_1, P_2$ and the recall is denoted by $R_1$ and $R_2$:
begin
\begin{equation}
    \hat{m}_{Precision}= \frac{P_1+P_2}{2}
\end{equation}

\begin{equation}
    \hat{m}_{Recall} = \frac{R_1+R_2}{2}
\end{equation}


The macro-avg F1-score is the harmonic mean of the precision and recall. The weighted average ($\hat{w}$) takes into account the weights of individual classes while calculating overall metrics as shown below:
\begin{equation}
    \hat{w}_{Precision} = w_1P_1 + w_2P_2
\end{equation}

\begin{equation}
    \hat{w}_{Recall} = w_1R_1 + w_2R_2
    \end{equation}

where the weights ($w_1, w_2$) are the proportion of the class in the data set. The weighted average F1-score is the harmonic mean of the precision and recall.

\subsection{Experiment setup} and technical details 

\begin{figure*}[htbp!]
\centering
\includegraphics[width = 17.5 cm]{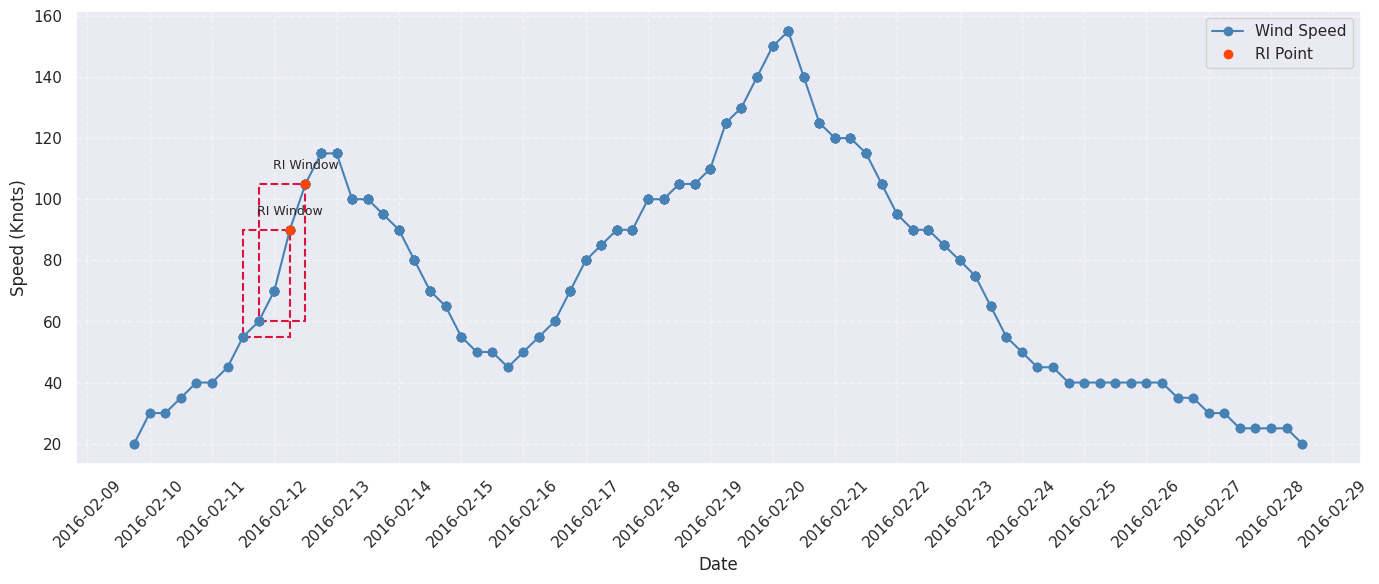}
\caption{ Life cycle of Winston cyclone in the South Pacific Ocean, highlighting a 24-hour window where RI event is detected (greater than or equal to 30 knots).}

\label{fig:ri-event}
\end{figure*}

 In the classification module given in Step 5 of Figure \ref{fig:framework}, the LSTM model input has $n$ data points representing $n \times 6$  hours.  We will evaluate the best value for $n$, experimentally.   The temporal cyclone data is sampled at regular 6-hour intervals. Our goal is to detect  RI events defined by a change of 30 knots in the wind intensity over 24 hours from a given data point as shown in Figure \ref{fig:ri-event}.

 
 We use the Adam optimiser \citep{kingma2014adam} as the designated training algorithm for all the LSTM-based models. We determined the hyperparameter values based on the data and trial experimental runs.  Table \ref{tab:paraSet} presents the hyperparameters for multivariate LSTM models (Model 2 in Figure \ref{fig:models})  used as a classification model (Step 5) and data augmentation module (Step 3) in the Framework (Figure \ref{fig:framework}.  We use the mean-squared error (MSE) loss for the data augmentation module. In the case of the Univariate LSTM (Model 1 in Figure \ref{fig:models}), we use 1 input neuron with the same setup as the rest of the hyperparameters. We implemented the Ensemble LSTM and Hybrid Ensemble LSTM by using the same setup as Table \ref{tab:paraSet} that defined the Multivariate and Univariate LSTM models.   We use 75 percent data in the training set and 25 percent data in the test set, where the data was separated based on period.  The training/test set was created for the respective oceans as shown in Table \ref{tab:data}. We ran 30 independent model training using random initial weights and biases of the respective models.


\begin{table}[htbp!]
\small
\centering
 \begin{tabular}{l  l   l   l   l}
 \hline
  \hline
Module & Hyperparameters & Type & Num. \\
 \hline
  \hline 
Classification  &   &  & \\

   &  Input Neurons & --  &  3\ \\ 
  &  Hidden Neurons & ReLu  &  50\ \\ 
   
&  Output Neurons & Softmax & 2\\\  
& Training & Epochs & 100 \\  
&Loss function & Cross-entropy & --\\ 
 \hline 
Data Augmentation &   &  \\

   &  Input Neurons & --  &  3\ \\ 
  &  Hidden Neurons & tanh and sigmoid  &  50\ \\ 
   
&  Output Neurons & Linear & 2\\\  
& Training & Epochs & 100 \\  
&Loss function & MSE & --\\ 
 \hline 
 \hline  

\end{tabular}
\caption{Hyperparameters for multivariate LSTM models (Model 2 in Figure \ref{fig:models})  used as classification model (Step 5) and data augmentation module (Step 3) in Framework (Figure \ref{fig:framework}. We use the mean-squared error (MSE) loss for the data augmentation module. In the case of the Univariate LSTM (Model 1 in Figure \ref{fig:models}), we use 1 input neuron with the same setup for the rest of the hyperparameters.}
\label{tab:paraSet}
\end{table}

 \section{Results} 

 \subsection{Preliminary Analysis and Data Visualisation }

 \begin{figure*}[htbp!] 
 
\centering     
\subfigure[South Indian Ocean]{\label{fig:a}\includegraphics[width=80mm]{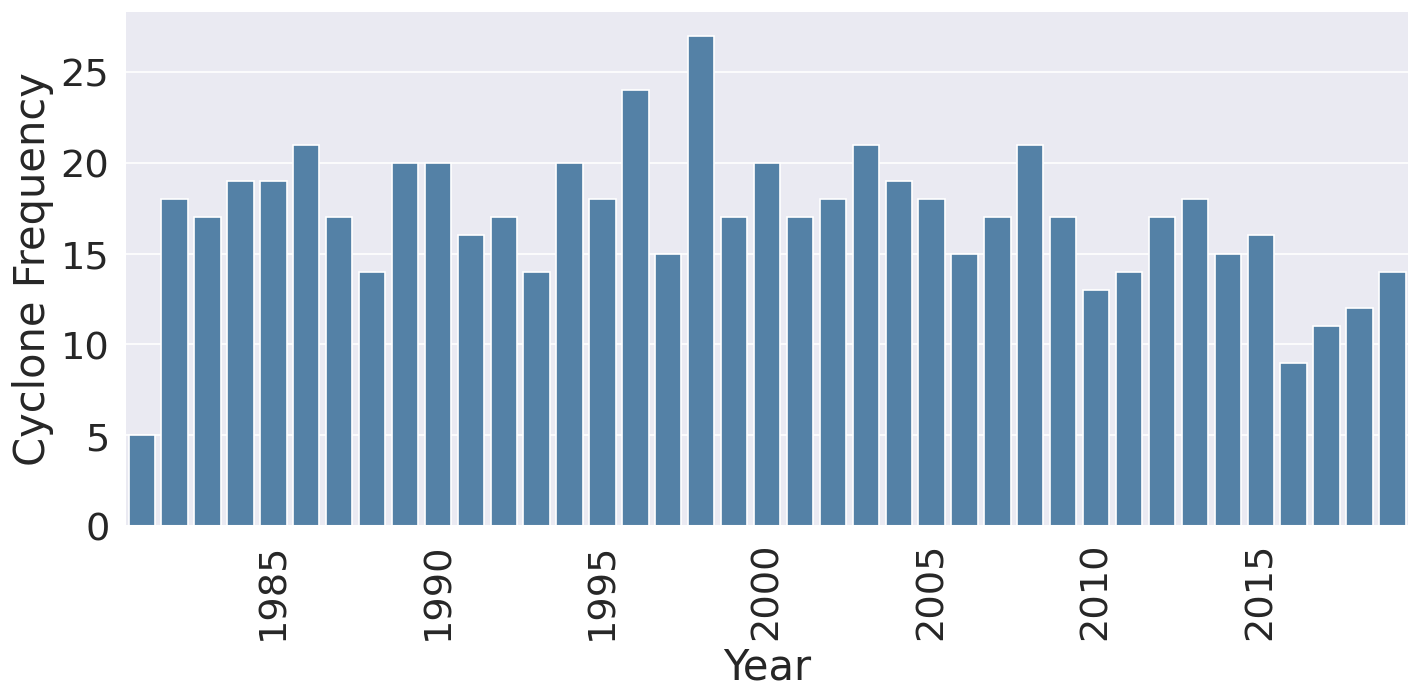}}
\subfigure[South Pacific Ocean]{\label{fig:b}\includegraphics[width=80mm]{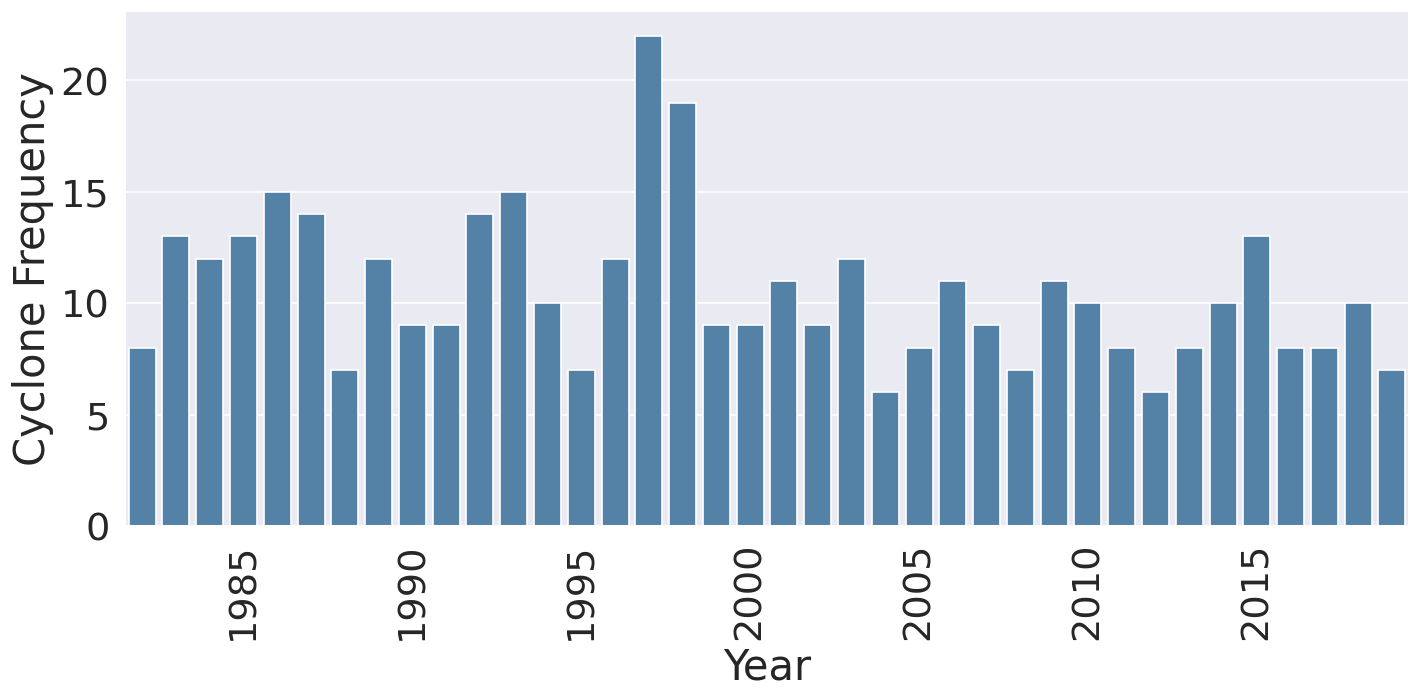}}

\caption{Cyclone frequency over the years for the respective datasets.}
\label{fig:freq}
\end{figure*}

 \begin{figure*}[htbp!] 
 
\centering     
\subfigure[South Indian Ocean]{\label{fig:a}\includegraphics[width=80mm]{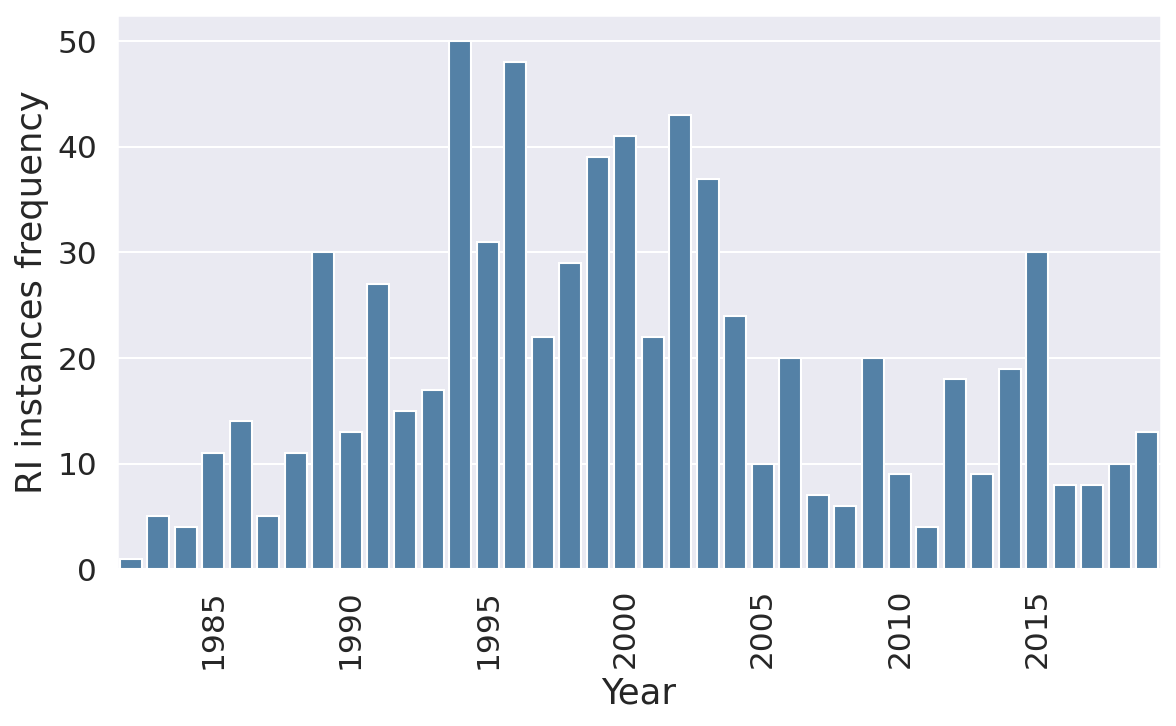}}
\subfigure[South Pacific Ocean]{\label{fig:b}\includegraphics[width=80mm]{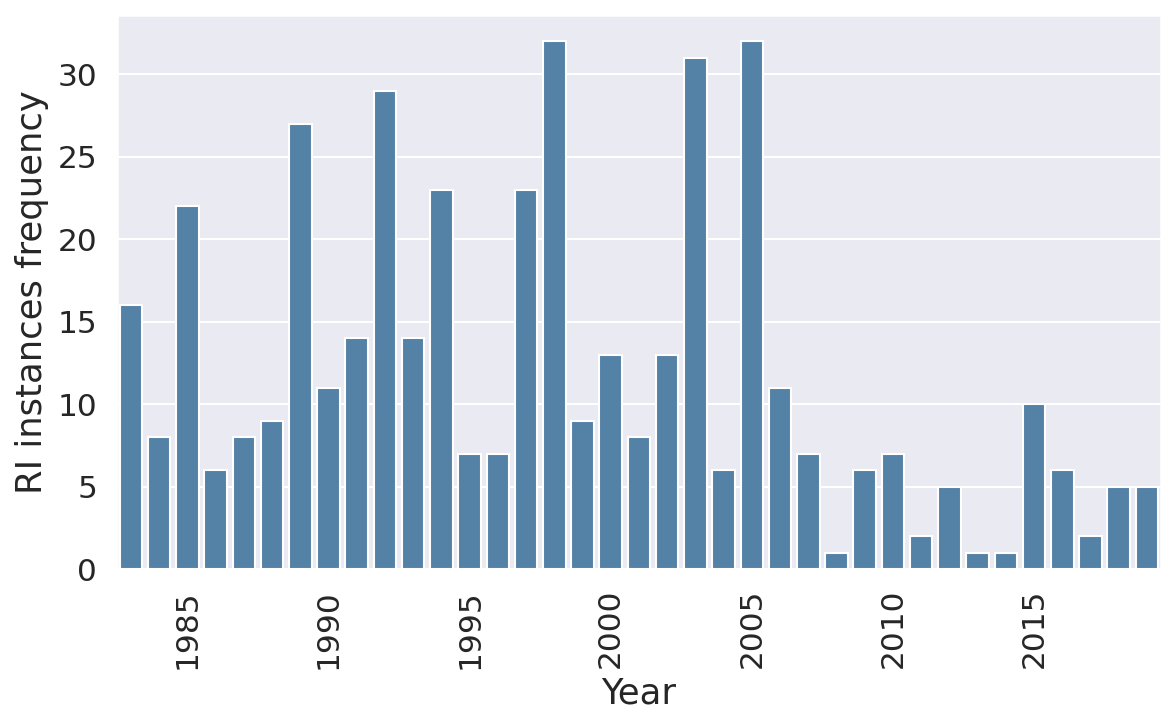}}

\caption{Frequency of RI events for the respective datasets.}
\label{fig:ri-freq}
\end{figure*}

\begin{figure}[htbp!]
\centering
\includegraphics[width = 8cm]{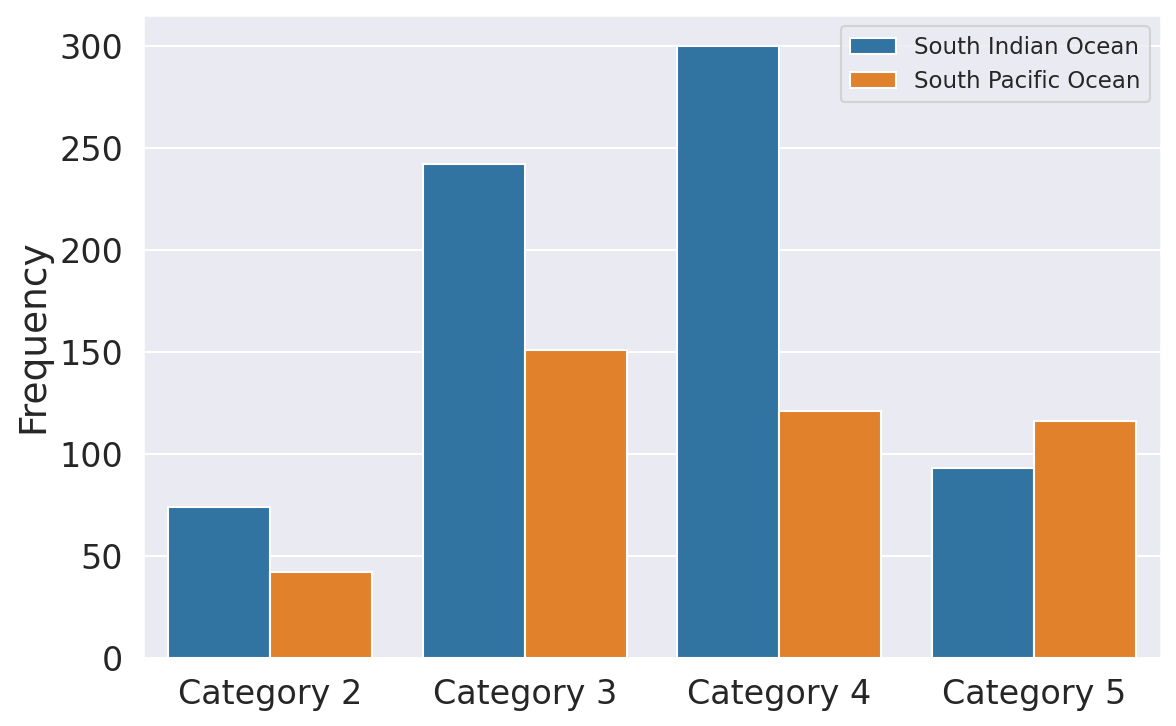}
\caption{Number of RI events within each cyclone category for the two oceans.}
\label{fig:categoryvsri}
\end{figure}


We first present a preliminary analysis to get an understanding of the data, taking into account the cyclone frequency, and RI cases for the timeframe (1980-2020) for both train and test data in Table \ref{tab:data}. 

Figure \ref{fig:freq} presents a visualisation regarding the cyclone frequency for the South Indian Ocean and South Pacific Ocean, respectively. In the case of the South Indian Ocean, we observe a significant change in the number of cyclones in 1996 and 1998, and in the case of the South Pacific Ocean, we observe a significant change in the years 1997 and 1998. Figure \ref{fig:ri-freq} presents the frequency of RI events for the respective ocean regions. We can observe that a larger number of RI cases occurred for cyclones from 1995 to 2004 in the South Indian Ocean, and there was a much lower number of RI cases afterwards. Furthermore, there is a much lower number of RI cases after 2008 for the South Pacific Ocean. Hence, we can generalise that there has been a much lower number of RI events in the last decade for both ocean regions. The difference in these decadal trends could be due to the change of climate patterns, such as the El Ni{\~n}o and La Ni{\~n} cycles \cite{meyers2007years,vecchi2010nino}.

Table \ref{tab:data} shows the percent of minority classes over the entire training and testing data, where the two classes refer to RI events and non-RI events. The instances in Table \ref{tab:data}  refer to sliding windows (elapsed time of 24 hours, among 6 data points).  We note that the portion for the minority class is very small. 
The South Indian Ocean has a larger number of cyclone events, when compared to the South Pacific Ocean. We observe that the South Indian Ocean has a higher class imbalance than the South Pacific Ocean. Furthermore, there is more training data with minority class for the South Pacific Ocean.

\begin{table*}[htbp!]
\small
\centering
 \begin{tabular}{l  l l  l l   l   l   l} 
 \hline
  \hline
 Dataset  & Num. Cyclones & Description      & Time-frame  &  Instances & Minority Class (\%) \\
 \hline
  \hline
 South Pacific Ocean & & Train & 1980-2020    & 7272 & 5.16
 \\ 
 & & Test  & 1980-2020   & 2424 & 1.15
 \\  
 \hline
 South Indian Ocean &  & Train & 1980-2020    & 15114 & 3.86 \\ 
&  & Test   & 1980-2020  & 5038  &  2.04
 \\  
  \hline  
   \hline
\end{tabular}
\caption{Dataset description showing the training and test split and time-frame, along with the number of instances, and percentage of RI cases (minority class).}
\label{tab:data}
\end{table*}




Figure \ref{fig:categoryvsri} presents the number of RI events given the different cyclone categories for the South Pacific (SP) and South Indian (SI) basins, respectively. In both cases, we observe that most of the RI events occur when the cyclone is in a category 3 phase and transitions into category 4. We note that RI events were not found in the early stages of cyclone genesis, i.e., category 1. The South Indian Ocean has a significantly larger number of RI events in Category 3 and 4 cyclones when compared to the South Pacific Ocean.


\subsection{Model evaluation}

\begin{table*}[htbp!]
\small
\centering
 \begin{tabular}{l l  l   l   l   l   l   l   l   l } 
 \hline
  \hline
 Strategy &   $n$  & Method &  Precision & Recall & F1-score \\
 \hline
Non-RI & 5 &   U-LSTM & 0.9920±0.0006  & 0.9942±0.0017  & 0.9931±0.0006
 \\ 
& 5 & M-LSTM & 0.9900±0.001  & 0.9828±0.0017  & 0.9864±0.0005
\\
 \hline
 
RI & 5 &  U-LSTM & 0.4422±0.0442  & 0.3548±0.05  & 0.3886±0.0279

\\
& 5 & M-LSTM &  0.3886±0.014  & 0.5245±0.0468  & 0.4455±0.0204
  \\

 \hline 
Non RI & 6 & U-LSTM & 0.9884±0.0013  & 0.9918±0.0023  & 0.9901±0.0006
 \\
 & 6 & M-LSTM & 0.9867±0.0009  & 0.994±0.0018  & 0.9903±0.0006
 \\ 
  \hline
RI & 6 & U-LSTM &0.536±0.0352  & 0.4427±0.0634  & 0.4798±0.0269
 \\ 
& 6 & M-LSTM & 0.5663±0.0582  & 0.3573±0.0453  & 0.4336±0.0294\\

\hline
Non RI & 7 & U-LSTM &  0.9948±0.0457  & 0.99±0.0578 & 0.9924±0.0478 \\
 
& 7 & M-LSTM &  0.9927±0.0226  & 0.997±0.0453  & 0.9948±0.0345 \\
RI & 7 & U-LSTM & 0.3784±0.0768  & 0.5385±0.0345  & 0.4444±0.0456  
 \\ 
& 7 & M-LSTM & 0.5625±0.0357  & 0.3462±0.0678  & 0.4286±0.0278  \\

\hline
Non RI & 8 & U-LSTM &  0.9932±0.0014  & 0.9897±0.0031  & 0.9915±0.0012
 \\ 
& 8 & M-LSTM &  0.9932±0.0007  & 0.9889±0.0038  & 0.991±0.0018 \\
RI & 8 & U-LSTM &   0.3087±0.0623  & 0.404±0.1245  & 0.3438±0.0707 
 \\ 
& 8 & M-LSTM &    0.3029±0.0671  & 0.4±0.062  & 0.339±0.0531
\\ 
\hline  
\end{tabular}
\caption{Results for South Indian Ocean (precision, recall and F1-score) for predicting cyclone RI cases using  Univariate and Multivariate models (U-LSTM and M-LSTM) for different model input timeframes ($n$). Note that both models are very poor in the detection of RI events due to class imbalance. }
\label{tab:modelinputSPO}
\end{table*}

 

\begin{table*}[htbp!]
\small
\centering
 \begin{tabular}{l l  l   l   l   l   l   l   l   l } 
 \hline
  \hline
 Strategy & Method &  Precision & Recall & F1-score \\
 \hline
Non-RI &   U-LSTM & 0.9927±0.0014  & 0.9903±0.0019  & 0.9915±0.0011

 \\ 
& M-LSTM & 0.9932±0.0007  & 0.9889±0.0038  & 0.991±0.0018
 \\ 
& E-LSTM  & 0.9938±0.001  & 0.9873±0.0047  & 0.9905±0.0028
 \\ 
& HE-LSTM  & 0.9935±0.001  & 0.9897±0.0051  & 0.9916±0.0024
 \\ 
 \hline
 
RI & U-LSTM & 0.3087±0.0778  & 0.3762±0.1208  & 0.3368±0.0927

\\
& M-LSTM & 0.3029±0.0671  & 0.4±0.062  & 0.339±0.0531
  \\
& E-LSTM & 0.3142±0.1249  & 0.456±0.0898  & 0.3669±0.1132
  \\
& HE-LSTM & 0.3539±0.1103  & 0.424±0.0898  & 0.3701±0.0681
  \\
 \hline 
Macro-average&U-LSTM & 0.6507±0.0395  & 0.6832±0.0603  & 0.6641±0.0467

 \\
(RI and Non-RI)  & M-LSTM & 0.648±0.0336  & 0.6944±0.0303  & 0.665±0.0272
 \\ 
& E-LSTM & 0.654±0.0629  & 0.7216±0.0466  & 0.6787±0.058
\\
& HE-LSTM & 0.6737±0.0551  & 0.7069±0.0438  & 0.6808±0.0349
 \\ 
  \hline
Weighted-average & U-LSTM & 0.9848±0.0022  & 0.9832±0.0022  & 0.9839±0.002

 \\ 
(RI and Non-RI)  & M-LSTM & 0.9855±0.0011  & 0.9823±0.0036  & 0.9837±0.0023
\\ 
& E-LSTM & 0.9862±0.0023  & 0.9814±0.0054  & 0.9836±0.004
\\ 
& HE-LSTM & 0.9863±0.0015  & 0.9834±0.0047  & 0.9846±0.003
\\ 
  \hline  
\end{tabular}
\caption{Results for South Pacific Ocean (precision, recall and F1-score) for predicting cyclone RI cases using univariate (U-LSTM), multivariate (M-LSTM), ensemble (E-LSTM) and hybrid models (HE-LSTM).}
\label{tab:modelsSPO}
\end{table*}

\begin{table*}[htbp!]
\small
\centering
 \begin{tabular}{l l  l   l   l   l   l   l   l   l} 
 \hline
  \hline
 Strategy  & Method &  Precision & Recall & F1-score\\
 \hline 
 Non-RI & U-LSTM & 0.9892±0.0011  & 0.9915±0.0026  & 0.9903±0.0009
 \\ 
& M-LSTM & 0.9889±0.0005  & 0.9918±0.0021  & 0.9903±0.001
 \\ 
 & E-LSTM  & 0.9885±0.0011  & 0.9914±0.0028  & 0.99±0.0009
 \\
& HE-LSTM & 0.9899±0.0009  & 0.9882±0.0018  & 0.9891±0.0011
 \\   
 \hline
RI &  U-LSTM & 0.5245±0.0543  & 0.4554±0.0549  & 0.4825±0.0257
\\
 & M-LSTM & 0.5439±0.0679  & 0.4402±0.0249  & 0.4866±0.0289
\\
& E-LSTM & 0.5074±0.0644  & 0.4196±0.0598  & 0.4519±0.0275
  \\
 & HE-LSTM & 0.4568±0.0496  & 0.4924±0.0474  & 0.473±0.0435
  \\

 \hline
Macro-average & U-LSTM & 0.7568±0.0268  & 0.7234±0.0263  & 0.7364±0.0129
 \\
 & M-LSTM & 0.7564±0.0339  & 0.716±0.0122  & 0.7331±0.0149
 \\ 
& E-LSTM & 0.7479±0.0318  & 0.7055±0.0287  & 0.7209±0.0137
 \\ 
  & HE-LSTM & 0.7234±0.0251  & 0.7403±0.0239  & 0.731±0.0223
 \\ 
 \hline
 
Weighted-average & U-LSTM & 0.9801±0.0009  & 0.981±0.0018  & 0.9804±0.0011
 \\ 
& M-LSTM & 0.9798±0.0013  & 0.981±0.002  & 0.9803±0.0015
\\ 
& E-LSTM & 0.9791±0.0008  & 0.9803±0.0017  & 0.9795±0.0009
\\ 
& HE-LSTM & 0.9795±0.0017  & 0.9786±0.0022  & 0.979±0.0019
\\ 
  \hline
\end{tabular}
\caption{Results for South Indian Ocean (precision, recall and F1-score) for predicting cyclone RI cases using univariate (U-LSTM), multivariate (M-LSTM), ensemble (E-LSTM) and hybrid models (HE-LSTM).}
\label{tab:modelsSIO}
\end{table*}

We first evaluate the results for the South Indian Ocean (precision, recall and F1-score) for predicting cyclone RI cases as shown in Table \ref{tab:modelinputSPO} using  Univariate and Multivariate models (U-LSTM and M-LSTM) for different model input timeframes ($n$).  Note that in all experiments, we report the mean and  95 \% confidence interval for 30 independent experimental runs (model training) using random initial parameters, i.e. weights and biases of the respective LSTM-based models. Since Precision and Recall are used to compute the F1-score, we only review the F1-score column in our analysis. Our focus is on the detection (prediction) of RI events. We provide results for Non-RI events merely for documentation from here onwards.  We notice that that both models are very poor in the detection of RI events due to class imbalance and $n=6$ is found to be optimal considering the f1 score metric and length of the sequence, i.e. longer model input data is helpful for the detection of RI cases; therefore, we use $n=6$ here onwards.

We next present the results given by classification performance of the four benchmark models,  including univariate (U-LSTM), multivariate (M-LSTM), ensemble (E-LSTM) and hybrid models (HE-LSTM) given in Figure \ref{fig:models}.  Table \ref{tab:modelsSPO} and \ref{tab:modelsSIO} shows the performance metrics of the four benchmark models for the South Pacific and South Indian Oceans, respectively. We include   the Precision, Recall and the F1-score to ensure that we capture the class imbalance problem. We first present the metrics (Precision, Recall and F1-score)  for individual classes (RI and Non-RI) and for both classes as a whole (Macro-average and Weighted-average).  We note that the major goal is to predict the RI events, and the non-RI event metrics are just given as a baseline, along with the combined metrics (RI and non-RI).  The focus here is on review results that include RI  and the Macro-average F1-scores. In the case of the South Pacific Ocean (Table \ref{tab:modelsSPO}), we find the HE-LSTM provides the best F1-score score for detection of RI events. This is followed by E-LSTM and looking at the   Macro-average,  HE-LSTM is also the best. In the case of the South Indian Ocean (Table \ref{tab:modelsSIO}), where M-LSTM and U-LSTM are clear winners for the RI events, HE-LSTM provides close results when compared to them. Furthermore, in the case of Macro-average, these three models provide very similar results; clearly, E-LSTM performs the worst.


\subsection{Model Performance with Data Augmentation}

We next present the results using the LSTM-based data augmentation strategy (Figure \ref{fig:generate})  using M-LSTM since this was one of the  models showing good results for RI events in the four benchmark models for the South Indian Ocean.  

In Tables  \ref{tab:SP-DALSTM}, we repeat the M-LSTM results from Tables \ref{tab:modelsSPO} and \ref{tab:modelsSIO} so that it is easier to compare the results with the data augmentation counterpart DA-M-LSTM (M-LSTM trained on augmented dataset). In the case of the South Pacific Ocean (Table  \ref{tab:SP-DALSTM}), for detection of RI events, we observe that DA-M-LSTM, which features synthetic wind-intensity and tracks, provides the improved performance.  In the case of the South Indian Ocean (Table 6), we observe the same pattern for the detection of RI events. We did not include non-RI events for analysis alone, as this did not feature data augmentation. Furthermore, Macro-average and weighted average results also show that DA-M-LSTM  provides improved results. We note that the South Pacific  Ocean test dataset has much lower RI cases (Table 2) when compared to the South Indian Ocean, which may influence the Macro-average and Weighted-Average results. However, our main strategy is to access the results for detention RI events only and the rest of the results (Macro-average and Weighted-average) are merely for benchmarking future models. Next, we provide further analysis using DA-M-LSTM which provided the best results in our analysis.



\begin{table*}[htbp!]
\small
\centering
 \begin{tabular}{l l l   l   l   l   l   l   l   l } 
 \hline
  \hline
 Model & Strategy & Ocean &  Precision & Recall & F1-score \\

 \hline

M-LSTM & RI  & South Pacific & 0.3029±0.0671 & 0.4±0.062 & 0.339±0.0531

 \\
& & South Indian & 0.5439±0.0679  & 0.4402±0.0249  & 0.4866±0.0289
\\
 \hline

DA-M-LSTM & RI & South Pacific & 0.6937±0.0437  & 0.619±0.0594  & 0.6513±0.0355

 \\
& & South Indian & 0.8423±0.0454  & 0.5419±0.0567  & 0.657±0.0444

\\
\hline 
  
M-LSTM & Macro-average & South Pacific & 0.648±0.0336  & 0.6944±0.0303  & 0.665±0.0272

  \\
& & South Indian & 0.7564±0.0339  & 0.716±0.0122  & 0.7331±0.0149

\\
  \hline  

  
DA-M-LSTM & Macro-average & South Pacific & 0.8386±0.0215  & 0.8034±0.0287  & 0.8185±0.0182

  \\
& &  South Indian & 0.9122±0.0226  & 0.7689±0.0281  & 0.823±0.0227

\\
  \hline  
\end{tabular}
\caption{Comparison of Results for different oceans showing the Precision, Recall and F1-score for predicting RI and Non-RI  cases between data augmentation (DA) for M-LSTM and M-LSTM}
\label{tab:SP-DALSTM}
\end{table*}

\begin{figure*}[htbp!]
    \centering
    \subfigure[South Pacific Ocean]{
        \includegraphics[width=0.85\textwidth]{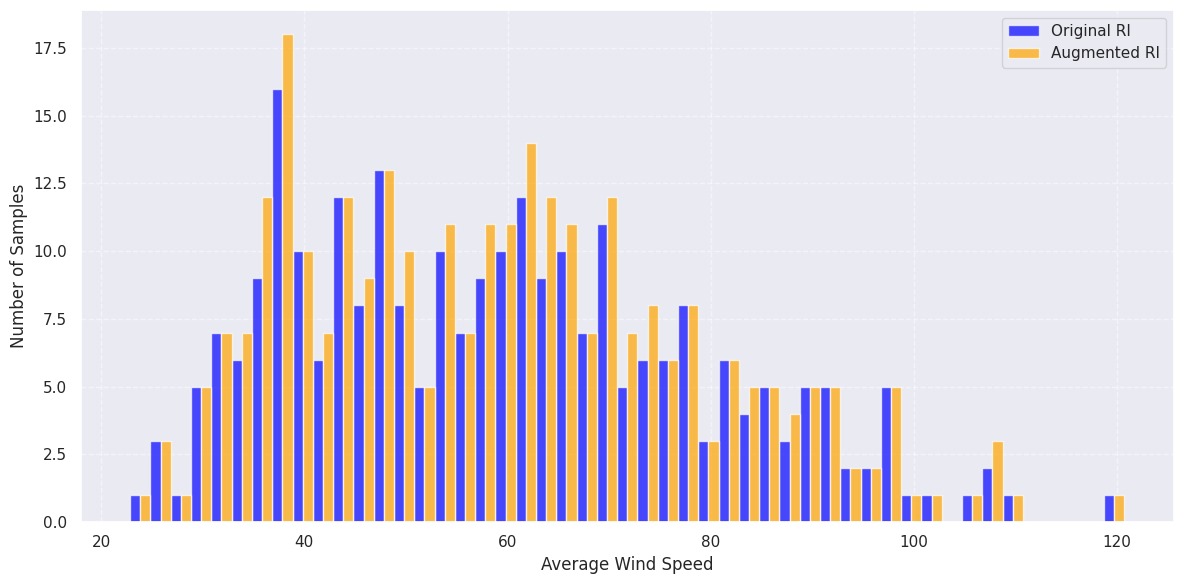}
        \label{fig:image1}
    }
    \hfill
    \subfigure[South Indian Ocean]{
        \includegraphics[width=0.85\textwidth]{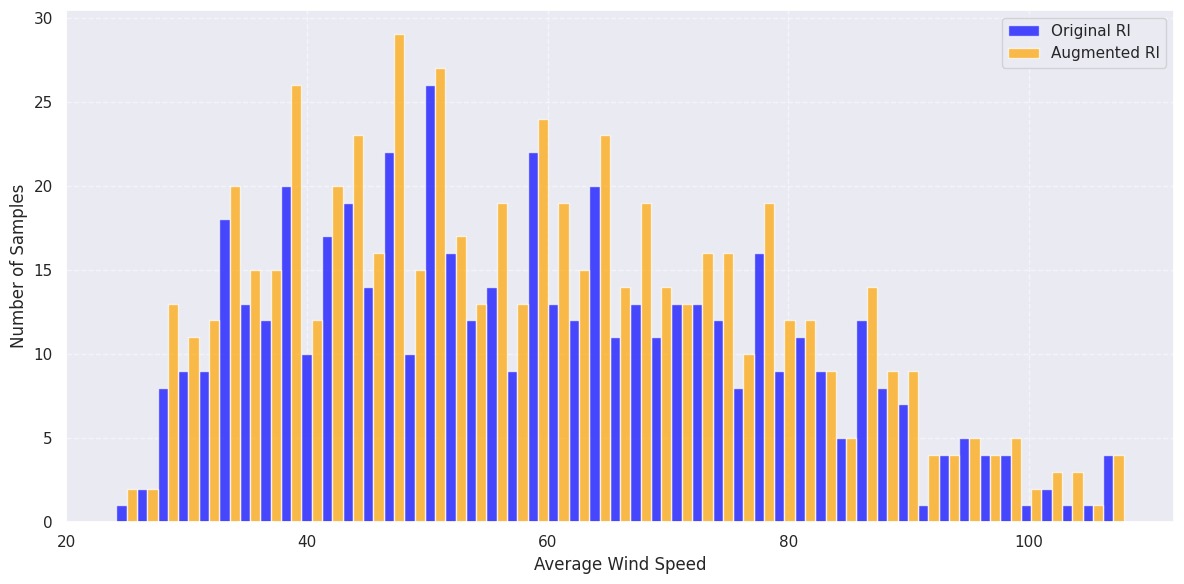}
        \label{fig:image2}
    }
    \caption{Histograms comparing the average wind intensity of original and augmented Rapid Intensification (RI) cyclone instances in the South Pacific and South
Indian Oceans}
    \label{fig:side_by_side}
\end{figure*}



\begin{figure*}[htbp]
\centering
\includegraphics[width = \linewidth]{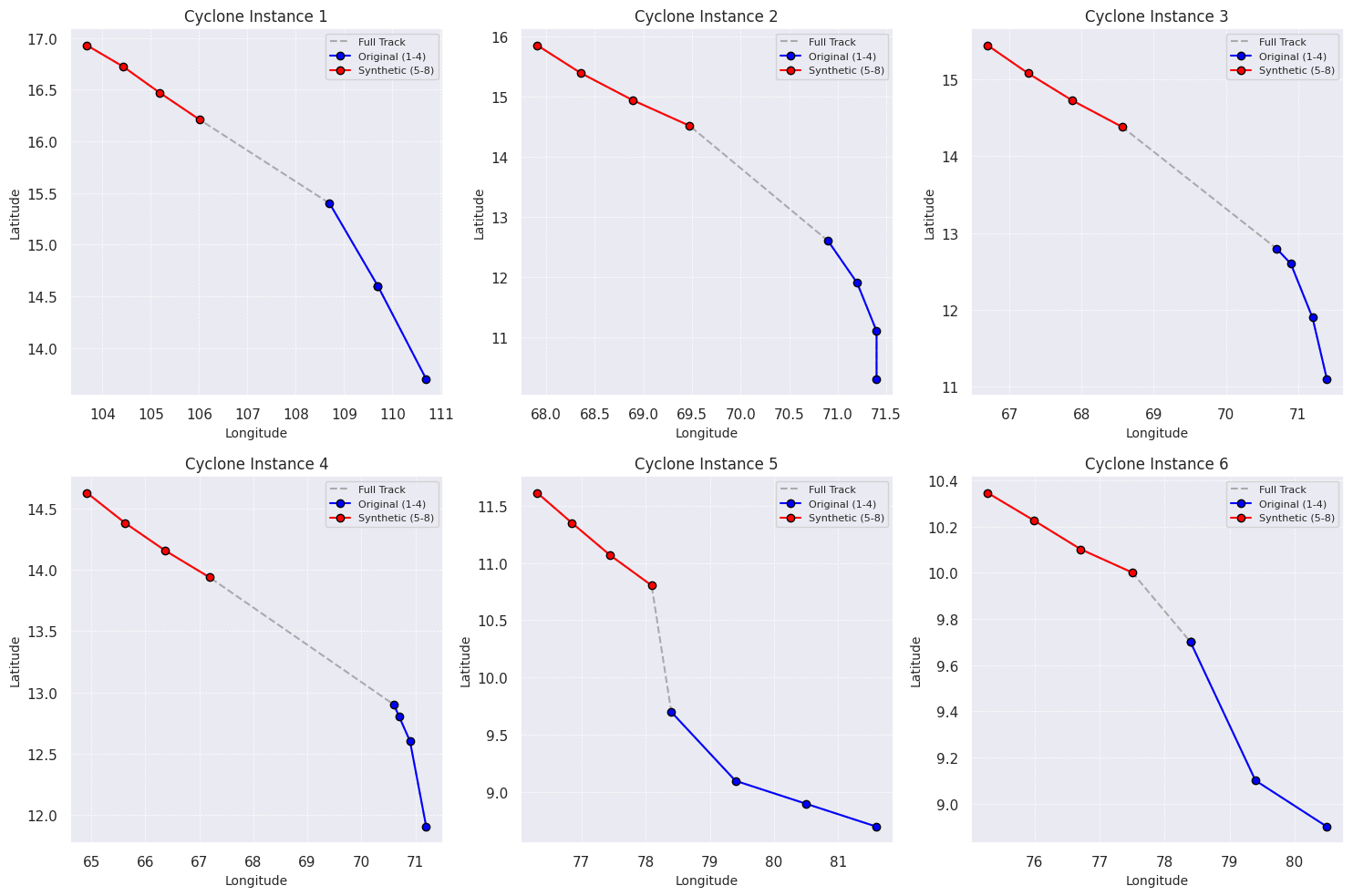}
\caption{Plots of tracks of synthetically generated cyclone instances}

\label{fig:syn}
\end{figure*}

Figure \ref{fig:syn} presents the tracks of synthetically generated cyclone instances, which were added to the training dataset for data augmentation. The first 4 points are taken from the original cyclone and the next 4 data points are synthetically generatd using LSTM model to predict (generate) wind intensity and augment the data of RI instances in the training dataset.

 


\begin{figure*}[htbp]
\centering
\includegraphics[width = \linewidth]{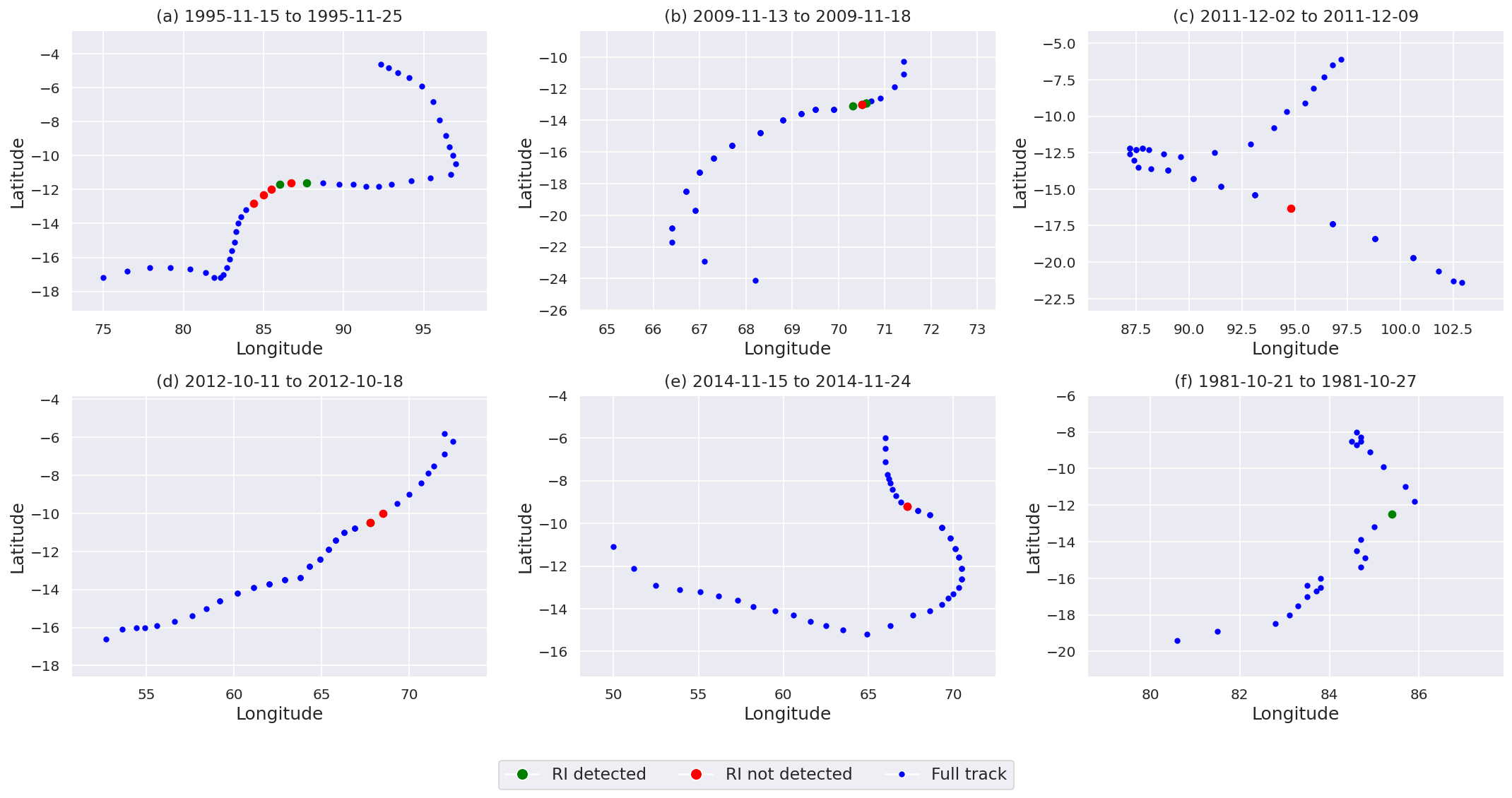}
\caption{Cyclone tracks showing RI points detected and those missed by DA-M-LSTM}

\label{fig:data}
\end{figure*}

 \section{Discussion}





Climate-related extreme events present class imbalance problems due to an  anomalous concentration of extreme events due to global warming \cite{mora2018extreme}. However, our analysis shows that in the case of cyclones, the increasing trend is not so clear in the case of RI events, since there were more RI events pre-2010 era (Figure 7). Cyclone formation depends on sea-surface temperatures, and we note that there have been changes in high-category cyclones in past decades. The chaotic nature of cyclones makes it difficult to represent cyclone characteristics such as rapid intensification with a simple binary value (RI vs non-RI) \citep{faghmous2014big} which presents a continuous challenge for modelling. It is clear that RI events are more observable in Category 3 and 4 cyclones (Figure 8), taking into account the data of the past decades. 

 Although our study presents an effective deep learning framework for prediction of RI events in cyclones, it is crucial to acknowledge certain limitations. Notably, the exclusion of key meteorological features, such as air temperature and air pressure of the cyclone, poses a constraint on our predictive model. Air temperature and air pressure are fundamental factors influencing cyclone dynamics \cite{chen2019pathway}, and their omission from our analysis may impact the nuanced understanding of the atmospheric conditions leading to RI events. These factors could provide  valuable insights into the intricate processes governing cyclonic intensification, and their incorporation could refine the accuracy and reliability of our model predictions. However, our current data source does not provide this information, and we need to acquire data from other courses to further enhance our model. It would also be possible that data would not be available in a coherent manner, and missing data may pose further challenges, which can be addressed by data augmentation methodologies as given in our framework. Rather than generating synthetic tracks and wind-intensity, the framework could be extended to generate air temperature data, given that such data is available for some of the cyclones.As we chart the path forward, it becomes evident that future research endeavors should prioritize the integration of air temperature and air pressure data into the predictive framework.Moreover, this study establishes quantitative performance benchmarks across multiple deep learning architectures for the task of cyclone RI detection. These benchmarks, based on precision, recall, and F1-score metrics, provide a reproducible and comparative foundation for future research aimed at improving detection accuracy or exploring alternative model architectures and data modalities.



In the literature, there has been much focus on GAN and SMOTE-based methods for class imbalanced problems. Recently, Khan et al. \cite{Khan2024} provided a review about their  combination with ensemble models; however, this was not suitable for our problem as we are dealing with spatiotemporal problem, not tabular data. 
Our study deviates from other applications of GANs in cyclone-based cloud image generation \cite{xu2019cyclone} by focusing solely on wind-intensity data for the detection of rapid intensification. Combining SMOTE with GAN, as done in prior pattern classification studies \citep{sharma2022smotified}, underscores the class imbalance challenges prevalent in climate extremes, especially within the context of cyclone prediction. In acknowledgment of the limitations associated with SMOTE and GANs, particularly their inability to predict spatiotemporal coordinates, we opted for a different strategy. Instead of relying on these methods, we employed the LSTM model to introduce data for cyclone data (spatial coordinates and wind intensity). This approach facilitated oversampling of RI events while preserving the inherent spatiotemporal characteristics of the data.

 GANs face challenges when the generated distribution does not substantially overlap with the target data distribution. The generator's ability to create arbitrary data points may influence training performance. Configuring hyperparameters, such as learning rate and updates to generator and discriminator components, becomes crucial in addressing these challenges. Ongoing efforts to enhance GANs, including the development of variants like Wasserstein GAN, reflect the commitment to overcoming limitations in data augmentation \citep{goodfellow2017nips, arjovsky2017wasserstein, li2018limitations, gui2021review}. Developing GAN variants for data augmentation in the case of spatio-temporal data can be done in future research, which may provide additional insights into improving the accuracy of predictions.

 We note that there are also avenues to strengthen the benchmark models in this study further, which can improve RI detection with data augmentation. Xu et al. \cite{xu2023tfg} recently used graph CNNs for cyclone intensity prediction, which extracted general spatial features, subtle spatial features, and general wind speed information from temporal data. This approach opens the path to incorporate the framework in our study for the prediction of  RI events. Furthermore, our framework can be improved by utilising uncertainty quantification in predictions using Bayesian deep learning methods \cite{Kapoor2023,chandra2021bayesian,chandra2023bayesian}

Future work can include implementing our models by adding important features like air temperature, air pressure, crucial for rapid intensification. 

 \section{Conclusion}
 
 In this study, we presented a deep  learning framework   to tackle the challenge of class imbalance in the prediction of RI events in cyclones. Our study employed a novel strategy utilizing four distinct models that utilised univariate and multivariate LSTM models, taking into account combinations of wind-intensity and cyclone trajectory data.

Our results  indicate that adding spatial coordinates (trajectory) of cyclones provides better accuracy using benchmark multivariate LSTM models. Furthermore, we used a multivariate LSTM model to generate synthetic cyclone data to address limited cyclone RI events. We combined the synthetic data with real data in the  data augmentation framework  using an LSTM model for prediction of RI events which demonstrated improvement in accuracy of RI event prediction. 

The complex nature of the factors contributing to rapid intensification underscores the existing gaps in our understanding. Our study, though limited to the trend of wind intensity, highlights the necessity for a meticulous feature selection process to unravel the intricate dynamics leading to RI events. We note that the length of observations in the past had a major effect on the prediction of RI events. A short span of data resulted in inferior results. 

In terms of practical implications, our research has a direct impact on meteorological organisations engaged in cyclone prediction. By leveraging the ensemble of LSTM models, we present a more nuanced and diverse approach. This shift enables better disaster management, especially in the face of high-category cyclones, where RI events can lead to devastating consequences, including structural damage and the generation of massive wave surges resulting in shoreline flooding.

Hence, our research contributes not only to the advancement of cyclone prediction methodologies but also to the broader discourse on the application of machine learning in meteorology. The deployment of our diverse LSTM models offers a promising avenue for future research, encouraging a deeper exploration of cyclonic features and a more holistic understanding of the mechanisms governing rapid intensification.



  \section*{Code and Data Availability}

 Our framework implemented in Python is available via Github repository with code and data:
 \footnote{ \url{https://github.com/sydney-machine-learning/cyclone\_deeplearning}} 
 
  \section*{Author Credit Statement }
    
 V. Sutar and A. Singh contributed to programming, experiments and writing. R. Chandra conceptualised and supervised the project and contributed to experiment design, writing and analyses.

  \section*{Conflicts of interest}
  
 I declare that the authors have no competing interests.

\bibliographystyle{num}

 \bibliography{GAN,cyclone}

\end{document}